\documentclass{article}

\usepackage{arxiv}

\usepackage[utf8]{inputenc} 
\usepackage[T1]{fontenc}    
\usepackage{hyperref}       
\usepackage{url}            
\usepackage{booktabs}       
\usepackage{amsfonts}       
\usepackage{nicefrac}       
\usepackage{microtype}      
\usepackage{lipsum}
\usepackage{graphicx}
 \usepackage{amsmath}
 \usepackage{xcolor}
\graphicspath{ {./images/} }

\title{AttentiveGRU: Recurrent Spatio-Temporal Modeling for Advanced Radar-Based BEV Object Detection}

\author{
 Loveneet Saini \\
  TMDT\\
  University of Wuppertal\\
  Wuppertal, Germany \\
  \texttt{Loveneet.Saini@aptiv.com} \\
   \And
 Mirko Meuter \\
  Radar Machine Learning\\
  Aptiv\\
  Wuppertal, Germany \\
  \texttt{mirko.meuter@aptiv.com} \\
  \And
  Hasan Tercan, Tobias Meisen \\
  University of Wuppertal \\
  Wuppertal\\
  \texttt{\{tercan, meisen\}@uni-wuppertal.de} \\
}

\begin{document}
\maketitle
\begin{abstract}
Bird's-eye view (BEV) object detection has become important for advanced automotive 3D radar-based perception systems. However, the inherently sparse and non-deterministic nature of radar data limits the effectiveness of traditional single-frame BEV paradigms. In this paper, we addresses this limitation by introducing AttentiveGRU, a novel attention-based recurrent approach tailored for radar constraints, which extracts individualized spatio-temporal context for objects by dynamically identifying and fusing temporally correlated structures across present and memory states. By leveraging the consistency of object's latent representation over time, our approach exploits temporal relations to enrich feature representations for both stationary and moving objects, thereby enhancing detection performance and eliminating the need for externally providing or estimating any information about ego vehicle motion. Our experimental results on the public nuScenes dataset show a significant increase in mAP for the car category by 21\% over the best radar-only submission. Further evaluations on an additional dataset demonstrate notable improvements in object detection  capabilities, underscoring the applicability and effectiveness of our method.
\end{abstract}


\section{Introduction}
\label{sec:intro}
Perception systems in today's vehicles support a wide range of applications, such as automatic emergency braking (AEB) and automatic valet parking (AVP). Thereby, the standard industry approach is to use a multi-sensor suite of radar, lidar, and cameras to facilitate these complex applications. Radar, known for its superior long-range capabilities and effectiveness in adverse weather conditions, as well as its Doppler measurement feature, has become an integral part of these perception systems. The automotive 3D radar system processes reflections from targets through a signal processing chain that estimates parameters such as time-of-flight and arrival angle, and produces range-azimuth-Doppler (RAD) feature maps of the scene within the radar's field of view. Despite its lower resolution and sparser data compared to lidar and cameras, radar's ability to capture point reflections from objects provides unique object detection capabilities through distinct target signatures in the radar measurement space. 

Recent deep learning approaches to radar data have explored a variety of input formats. The state of the art includes models that process a single radar frame either in a point cloud format, emphasizing specific local features \cite{pan2023moving}, or in a Bird's Eye View (BEV) grid format, facilitating the natural extraction of local patterns \cite{major2019vehicle,zhang2021raddet}. Notably, due to emerging trend of fusing different sensors like camera and radar in common BEV space, the research in radar exclusive BEV based detection has become of significant importance as it helps in establishing stronger backbones in such multi-modal architectures.

Radar, as a low-resolution sensor, is inherently affected by sparsity issues \cite{zhou2020mmw,zhou2022towards}. To mitigate this, some single-frame approaches have employed the concatenation of multiple radar frames to generate a denser input \cite{svenningsson2021radar}. However, this method of aggregation is computationally demanding and fails to leverage the advantages of exploiting temporal relationships. Hence, in this work, we propose a model that harnesses the temporal relationship patterns in BEV radar data using recurrent methods.

For denser input modalities like lidar, recent research underscores the advantages of incorporating temporal relationships. These multi-frame approaches \cite{calvo2023timepillars, hou2024query} generally begin by aligning pixel-wise BEV features from historical data with current inputs to compensate for ego-vehicle motion, utilizing either a learned or externally provided ego-pose matrix, and subsequently integrate them using ConvGRU \cite{ballas2016delving} or attention mechanisms \cite{vaswani2017attention}. A similar and more up-to-date approach is utilized for camera images, where historic information is pre-aligned using ego-vehicle information \cite{li2022bevformer, huang2022bevdet4d}. Specifically in \cite{li2022bevformer}, Li et al. have noted that the availability of an ego pose matrix allows the alignment of only stationary objects due to the varying dynamics of moving objects in the scene. Following such pre-alignment with external ego information, Li et al. have proposed using deformable attention to model temporal relations by concatenating features from past and present 2D BEV queries. However, these lidar and camera specialized methods are less robust to the constraints posed by commercial radar data (demonstrated below) unless high-resolution radar imaging techniques are employed, which comes with increased pre-processing, data requirements, and costs. 

\begin{figure}
	\centerline{\includegraphics[scale=0.8]{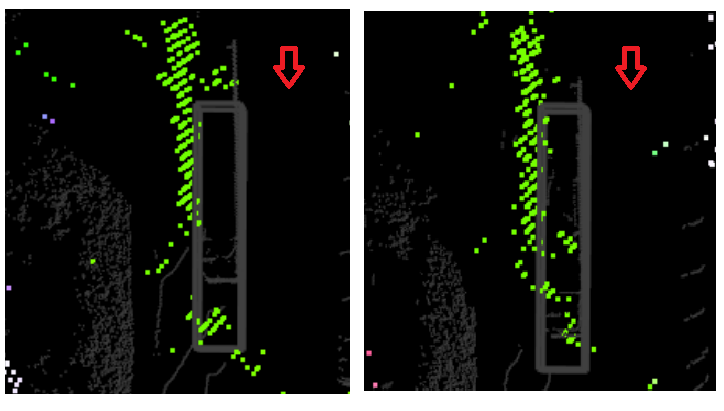}}
	\caption{Radar reflections for two consecutive scans from a truck approaching towards ego vehicle in the direction of arrow}
	\label{radar}
\end{figure}

In the radar measurement space, objects are detected by capturing point reflections of radar signal from objects (see Figure \ref{radar}) . Due to longer wavelengths of the radar sensor \cite{zhou2020mmw,zhou2022towards} the representations of objects in this space can be highly sparse with high dynamic noise. Specifically, the highly sparse reflections from the same object across consecutive scanned frames may vary in terms of the part of the object they originate from (front of truck in Figure \ref{radar}) , rendering them non-deterministic. Furthermore, unlike camera and lidar, the radar receiver antenna exhibits the highest sensitivity at the center of its beam (boresight), with sensitivity diminishing at greater angles. This attribute leads to progressively weaker observability of objects as they move away from the boresight. Consequently, depending on the dynamics of the object, its representation may appear smeared and significantly noisy when observed across consecutive frames.

The above mentioned challenges associated with the radar feature maps makes the direct use of temporal fusion methods from lidar and camera non-trivial. However, radar's ability to provide depth (range) and velocity (Doppler) of targets provides a unique opportunity to correlate objects over time. In typical driving scenarios, where objects do not abruptly change their motion, the range-azimuth features facilitate object matching across spatial depths, while the velocity feature aids in tracking moving objects across adjacent scans. In this paper, we present a novel plug and play temporal fusion layer for improving existing radar based object detectors for the widely used Range-Azimuth-Doppler (RAD) radar input format. Our proposed layer has linear computational complexity and is designed to improve object detection performance without the need for providing or estimating  ego motion based transformations. Specifically, our model incorporates a module inspired by the Perceiver \cite{jaegle2021perceiver}, which exploits the consistency of an object's latent encodings (query) across time frames to identify and fuse correlated object structures in a recurrent manner. We benchmark our proposed model using the public nuScenes dataset \cite{caesar2020nuscenes} and additionally use a larger and more complex proprietary radar dataset from \cite{braun2021quantification} for rigorous evaluation and detailed ablation studies. In summary, the contributions of this paper are as follows:

\begin{itemize}

\item We propose a temporal fusion layer tailored for radar data, to enhance detection performance by integrating individualized spatio-temporal contexts for each object recurrently over input time frames. This layer significantly reduces the computational effort by having linear complexity in both space and time and brings advantages of temporal fusion to highly sparse and challenging inputs.

\item We address a gap in the existing radar literature by combining the recurrence of GRUs with the attention mechanism to introduce a novel "attention gating" approach. This enhancement improves the ability of the GRU to selectively focus on temporally correlated locations for both moving and stationary objects across time frames, without the need for externally provided ego-motion information or explicit motion estimation.

\item We present an ablation study to analyze how global receptive fields and spatio-temporal representations influence detection performance for different object types, sizes and dynamics. This evaluation integrates the proposed layer with both traditional Feature Pyramid Networks (FPN) and modern transformer-based BEV object detection models.

\end{itemize}

\section{Related Work}
\label{sec:formatting}
\begin{figure*}[htbp]
    \centering
    \def\svgwidth{\linewidth}
    \fontsize{8}{10}\selectfont
    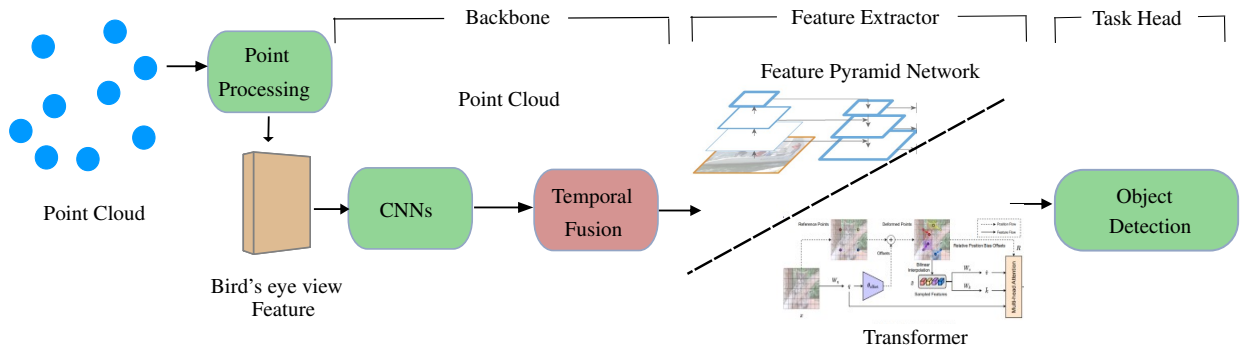
    \caption{Overall architecture with temporal fusion layer in the backbone}
    \label{overall}
\end{figure*}

\textbf{Multi-Frame object detection:}  Only a handful of recent works have addressed radar-only object detection in multi-frame scenarios. The studies by \cite{li2022exploiting} and \cite{yataka2024radar} model temporal relations by employing masked self-attention mechanisms between potential object centerpoints (referred as top K object features)  from consecutive time frame. As noted in their research, this method has limitations when applied to data from the nuScenes dataset, which uses commercial radar sensors, and is more appropriate for high-resolution radar images from advanced radar sensors \cite{sheeny2021radiate}. To overcome such limitations, we propose to perform temporal fusion using the latent space \cite{jaegle2021perceiver} of objects, instead of fusing over keypoints. By leveraging the consistency of an object's latent representation (query) across time frames, our method facilitates direct fusion over pixel-wise BEV features and brings the advantages of temporal fusion to highly sparse and challenging inputs. This is supported by observations that temporal fusion over pixel-wise BEV features is more beneficial for improving performance on sparser inputs than fusion over the keypoints.

A recent extension of \cite{li2022exploiting} and \cite{yataka2024radar} by Pu Wang et al. in \cite{yataka2024sira} explore temporal relations over a sequence of time steps rather than just consecutive frames. Their method models temporal relations using attention applied to potential object center points extracted from past T time steps. Specifically, they sample K subset features (extracted as K center points with heatmaps) from each step’s feature map using the same approach as in \cite{yataka2024radar} and group them into window patches. By exploiting the temporal order of the input sequence (termed "deformable temporal order"), these patches are reorganized into a new window that collects subset features from all past T time steps in temporal order. A masked self-attention mechanism is then used to update each subset feature. However, as the authors note, their approach is suitable for high-resolution radar image datasets and has limitations when applied to sparse nuScenes data \cite{li2022exploiting}. Furthermore, it has quadratic complexity with respect to the sequence length. In contrast, our approach directly fuses sampled BEV pixels instead of center points, allowing it to handle sparser inputs from commercial sensors more effectively than \cite{yataka2024sira}. In addition, by integrating recurrence with an attention mechanism, our method maintains linear computational complexity in both space and time. In contrast to fusion over keypoints, authors in \cite{decourt2024recurrent} have proposed the use of Convolutional LSTM layer\cite{shi2015convolutional} for temporal fusion by performing temporal fusion over different 2D views of 3D range-azimuth-Doppler (RAD) measurement space. However, as noted in \cite{calvo2023timepillars}, ConvLSTM is primarily effective only for stationary objects, since ego compensation is only able to align them across frames, by compensating for the movement of the ego vehicle but cannot align moving objects (each with different dynamics). 

In contrast to these radar-based approaches, our method eliminates the need to pre-align objects prior to temporal fusion, thereby avoiding explicit motion estimation effort for moving objects and even bypassing reliance on externally providing ego-motion information (using extra sensor) for stationary objects. Specifically, our approach dynamically identifies and samples temporally correlated locations for each object separately in both present and memory states. These locations are then recurrently fused via a state integration module, as described in Section 4.3.2. The learnable, deformable nature of this sampling mechanism across space and time allows the model to effectively adapt to the motion of both individual objects and the ego vehicle over time.

\section{Architecture Overview}

Our work extends the existing CNN and transformer based models for radar BEV object detection into the spatio-temporal domain to exploit temporal relationships from the RAD feature map of the scene within the radar's field of view. Traditionally, FPNs have been long used in many radar object detectors over the years \cite{ulrich2022improved, major2019vehicle, zhang2021raddet}, while only recently transformer based detectors have emerged \cite{Saini_2024_CVPR}. In order to utilize temporal relation, we first design a flexible architecture template which has a replaceable feature extractor to allow for substituting either FPN or a transformer. To this end, we adopt the popular Fully Convolutional One-Stage (FCOS) object detection framework \cite{tian2019fcos} as the basic template for our radar input as shown in Figure \ref{overall}. We adapt the original FCOS architecture such that it uses an input of pre-processed radar reflections by incorporating (i) a grid processing backbone with a temporal fusion module, (ii) a replaceable feature extractor, and (iii) a center point object detection head.

\subsection{Radar Signal Point-Processing}
\begin{figure}[htbp]
    \def\svgwidth{0.7\linewidth}
    \centering
    \fontsize{8}{10}\selectfont
    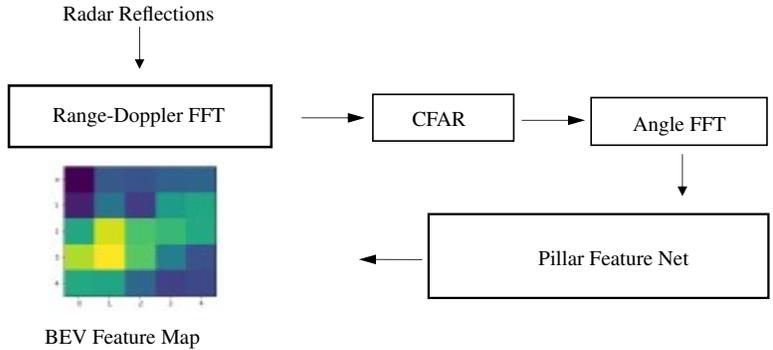
    \caption{Per-point processing chain for input radar reflections. Blue regions in the BEV map represent sparsity, while yellow is the peaks from reflections}
    \label{preprocess}
\end{figure}

Figure \ref{preprocess} illustrates the standard radar signal processing chain used to process the received reflections at each point. First, the radar signal undergoes a 2D Fast Fourier Transform (FFT), which facilitates the generation of a range Doppler spectrum. This spectrum is then processed through a Constant False Alarm Rate (CFAR) detector designed to effectively extract radar targets. Once detected, the arrival directions (angles) of the targets are estimated. This process is applied to all incoming radar point clouds to obtain common point features. The processed point cloud is then projected onto a 2D BEV grid, transforming it into an image-like input. For this projection, we use the Pillar Feature Net from the PointPillars methodology \cite{lang2019pointpillars}, which divides the cloud space into pillars similar to vertical bins. This approach is prevalent in much of the existing literature, such as \cite{K_hler_2023,xu2021rpfa,tan20223d,ulrich2022improved}, and is therefore adopted here "as is".

\subsection{Backbone}
In the backbone stage, the input BEV grid is first processed through convolution layers to introduce spatial awareness of the grid input into the network (see Figure \ref{overall}). The output sequence from these convolution layers is then downsampled by a factor of $K$ using a dynamic convolution module \cite{chen2020dynamic}. In order to improve the representational capacity for low-resolution objects in the input, this module has three parallel kernels $K= {1,2,4}$. A crucial component of this stage is the Temporal Fusion with AttentiveGRU module, which is responsible for extracting temporal relations. We will discuss this module in more detail in section \ref{sec:fuse}.

\subsection{Feature Extractor}

We designed our architecture to be flexible, allowing for a swappable feature extractor that can utilize either an FPN network or a transformer. Figure \ref{overall} illustrates the FPN-based feature extraction stage \cite{lin2017feature}, where the input is processed by multi-level CNNs to extract features at different resolutions, incorporating max-pooling and up-sampling layers. Additionally, as described, the FPN can be replaced by a transformer, as demonstrated in \cite{Saini_2024_CVPR}, enabling the use of a global receptive field that effectively captures global object features of varying sizes. 

\subsection{Head}
For the detection head, the target processor is inspired by the FCOS \cite{tian2019fcos} network, which uses a centerness score for keypoint estimation with heat map regression to identify and collect top-K center candidates. As previously demonstrated in Tian et al. research, FCOS is a suited alternative to anchor-based methods due to its capacity to utilise a multitude of foreground samples for the purpose of training the regressor. This makes it an optimal candidate for the detection head. The identified center candidates are finally regressed to obtain object properties after training using a focal loss \cite{lin2017focal}, similar to FCOS.

\section{Temporal Fusion with AttentiveGRU}
\label{sec:fuse}

\subsection{Overview}
\begin{figure*}[htbp]
    \centering
    \def\svgwidth{\linewidth}
    \fontsize{8}{10}\selectfont
    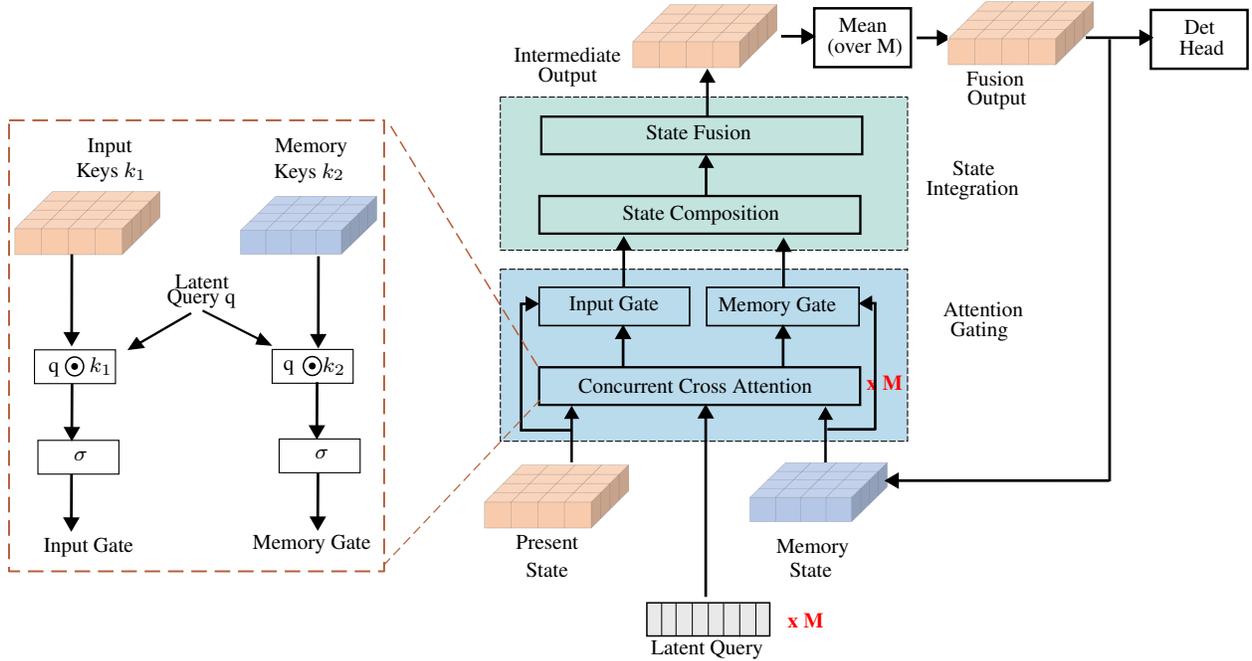
    \caption{Illustration of the proposed Temporal Fusion Block with Attention Gating (AG) and State Integration (SI) modules}
    \label{block}
\end{figure*}
Figure \ref{block} shows our proposed temporal fusion block, which consists of an attention gating and a state integration module. As shown in the figure, the block has three inputs, namely memory state (single BEV register buffer), present state (single BEV frame), and $M$ learnable latent queries (parameters). For an input sequence of length T, to the overall network (Figure \ref{overall}), each input frame is fed in a recurrent manner, starting from time step t=0 to t=T, as the current state, in the Figure \ref{block}. Thus, the entire input sequence passes through the fusion block in T passes. At each pass, except the last, the fusion output is used as an updated memory block, while the final T$^{th}$ pass fusion output is fed to the detector head. The goals of the block are first to dynamically learn temporally correlated locations across present and memory states for each object in the input using attention gating, and later to sample features from these locations and fuse them using the state integration module.

\subsection{Mechanism}

To illustrate the working principles of our fusion block, we detail its operation and theoretical underpinnings in terms of its individual modules as shown in Figure \ref{block}.  

\subsubsection{Attention Gating}
Each of the $M$ different learnable queries learn a latent-encoding for  $M$ different objects. For each object, the mechanism of detecting  correlated locations over time is based on the idea of existence of a unique latent encoding that remains consistent across time frames. This idea has its theoretical motivation in the work Perceiver \cite{jaegle2021perceiver}, in which Jaegle et al. proposed that an input modality (byte array) can be encoded into a latent representation (i.e., latent query), where each latent query serves as the centroid of clusters present in the input. Using visualization of attention maps, Jaegle et al. have extensively shown the feasibility of such encoding for diverse input images across the datasets and even other data modalities.

We extend this concept across multiple input time frames and propose that the latent representation query for each object (cluster) in the inputs ($M$ latent queries for $M$ different objects) is distinct and remains stable across time scans. Therefore, given an input of historical and present states, we can decode a learnable latent query to dynamically learn correlated pair of locations for each object over time, and then sample them to extract temporal features. Notably, the core assumption of our method will not hold in sudden acceleration and braking scenarios where conditions are rapidly changed. While our method is effective for typical driving scenarios, it may not generalize to out-of-distribution cases with such dynamic scenarios. 

Dynamically learning correlated pair of locations across time frames effectively addresses the non-deterministic and noisy nature of radar inputs, as the latent representation of the object remains consistent even when representations are smeared across adjacent frames.

As shown in the Figure \ref{block}, $M$ latent queries are used to dynamically learn locations over both historical memory state and present state for $M$ different objects. Specifically, a novel Concurrent Cross Attention operation is used where each of the  query $q$ jointly attends the present (keys $k_{1}$) and memory (keys $k_{2}$). On the resulting attention maps, a gating function $\sigma$ is applied to map attention scores to binary gates for dynamic selection of locations from present and memory. To make the dynamic selection process adaptive to the input, the gating function $\sigma$  is modeled as a learnable thresholding of the sigmoid layer. Essentially, a sigmoid layer processes the attention scores and the median of the sigmoid layer output is used as a learnable threshold resulting in binary gates. For radar data, the learned features of each of the latent query are higher dimensional embeddings of the RAD input to the overall network, so the dynamically learned locations are partially revealed using the spatial depth (range) and (Doppler) velocity information across frames.

The output of attention gating results in $2 \times M$ binary gates corresponding to each of the $M$ latent queries for the memory and present states, respectively. As the same latent query is used for both states, this facilitates the revelation of correlated locations through the corresponding gate. Consequently, at the input of the state integration module, each gate controls what information is passed on to the module by effectively populating the locations where the gate value is $1$ with features from the corresponding state. This operation results in a correlated pair of gated states from the memory ($h_{t-1}(q)$) and present state ($x_t(q)$)  for each latent query $q$.

\subsubsection{State Integration}
\begin{figure*}[htbp]
    \def\svgwidth{0.8\linewidth}
    \centering
    \fontsize{6}{8}\selectfont
    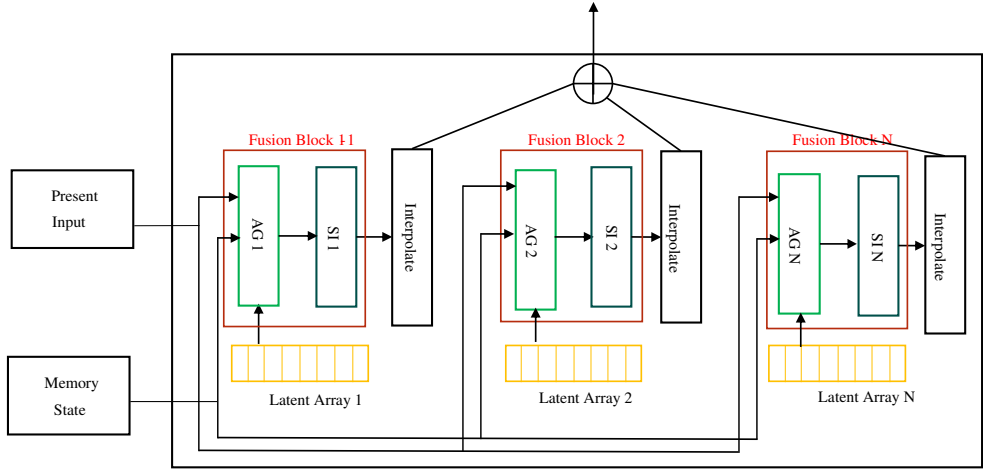
    \caption{An exemplary AttentiveGRU layer, showcasing the use of proposed Temporal fusion blocks with Attention gating (AG) and State Integration (SI) modules}
    \label{layer}
\end{figure*}
\begin{figure}[htbp]
    \def\svgwidth{0.7\linewidth}
    \centering
    \fontsize{8}{10}\selectfont
\begingroup%
  \makeatletter%
  \providecommand\color[2][]{%
    \errmessage{(Inkscape) Color is used for the text in Inkscape, but the package 'color.sty' is not loaded}%
    \renewcommand\color[2][]{}%
  }%
  \providecommand\transparent[1]{%
    \errmessage{(Inkscape) Transparency is used (non-zero) for the text in Inkscape, but the package 'transparent.sty' is not loaded}%
    \renewcommand\transparent[1]{}%
  }%
  \providecommand\rotatebox[2]{#2}%
  \newcommand*\fsize{\dimexpr\f@size pt\relax}%
  \newcommand*\lineheight[1]{\fontsize{\fsize}{#1\fsize}\selectfont}%
  \ifx\svgwidth\undefined%
    \setlength{\unitlength}{462.56014126bp}%
    \ifx\svgscale\undefined%
      \relax%
    \else%
      \setlength{\unitlength}{\unitlength * \real{\svgscale}}%
    \fi%
  \else%
    \setlength{\unitlength}{\svgwidth}%
  \fi%
  \global\let\svgwidth\undefined%
  \global\let\svgscale\undefined%
  \makeatother%
  \begin{picture}(1,0.52962809)%
    \lineheight{1}%
    \setlength\tabcolsep{0pt}%
    \put(0.23171448,0.0684687){\makebox(0,0)[t]{\lineheight{1.25}\smash{\begin{tabular}[t]{c}$\text{Overlapping}$\end{tabular}}}}%
    \put(0.2343327,0.01663907){\makebox(0,0)[t]{\lineheight{1.25}\smash{\begin{tabular}[t]{c}$\text{Areas}$\end{tabular}}}}%
    \put(0,0){\includegraphics[width=\unitlength,page=1]{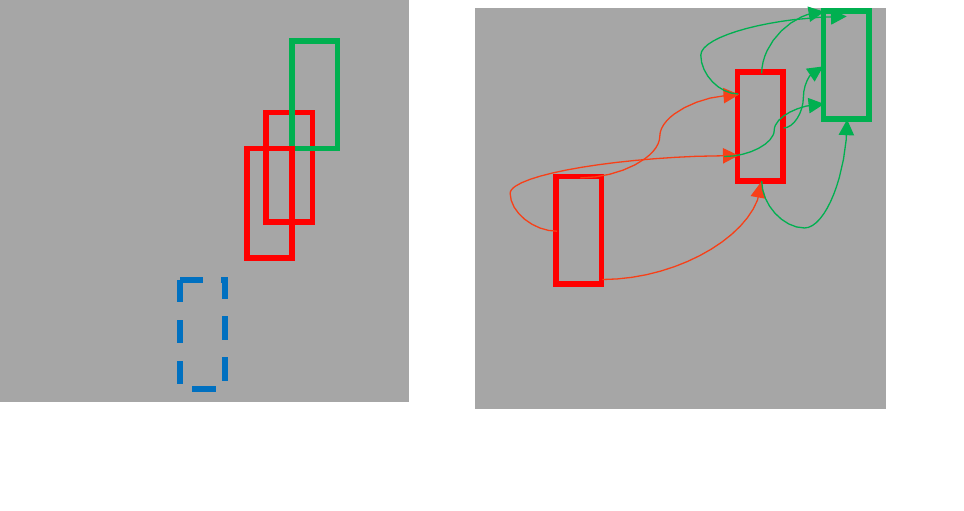}}%
    \put(0.71579011,0.0637017){\makebox(0,0)[t]{\lineheight{1.25}\smash{\begin{tabular}[t]{c}$\text{Non Overlapping}$\end{tabular}}}}%
    \put(0.60145793,0.03671922){\makebox(0,0)[t]{\lineheight{1.25}\smash{\begin{tabular}[t]{c}-\end{tabular}}}}%
    \put(0.72218352,0.01187207){\makebox(0,0)[t]{\lineheight{1.25}\smash{\begin{tabular}[t]{c}$\text{Areas}$\end{tabular}}}}%
    \put(0,0){\includegraphics[width=\unitlength,page=2]{deform.pdf}}%
  \end{picture}%
\endgroup%

    \caption{Visualization of composite state $C(q)$: Blue indicates the ego vehicle, red marks gated state from the memory ($h_{t-1}(q)$), and green represents the one from present ($x_t(q)$)}
    \label{combine}
\end{figure}

To extract temporal features from the resulting pairs of gated states, we modify the processing of vanilla ConvGRU \cite{ballas2016delving}. The vanilla version of the operator directly process the entire input by using reset and updates gates as modeled by equations 1-4 in the paper \cite{ballas2016delving}. Since, we have gated states for each object separately available, rather than the entire input, we reformulate the original equations to instead process our gated states ($h_{t-1}(q)$ and $x_t(q)$) for temporal features extraction. Specifically, for each pair of gated states we create a composite state $C(q)$ by concatenating gated states  ($\left[h_{t-1}(q), x_t(q)\right]$) for each of the M queries $q$ (called the state composition in Figure \ref{block}). Such composite state contains overlapping and non-overlapping regions of the same object across time in feature space, as shown in the Figure \ref{combine}. Hence, we can combine these areas to extract BEV temporal features for each object separately. For overlapping areas, we use local $1\times1$  convolution based gates $l_1(q)$  and $l_2(q)$ to obtain the fused state $h_t(q)$ as explained in equations \ref{eq}. In contrast to the global gating operation of attention gating module, the use of local gate facilitates sequential processing for each pixel in present and memory state and allows the extraction of temporal features that detail the object's local movement across its own spatial dimensions (left of Figure \ref{combine}). Based on overlapping features, gate $l_1(q)$ is used to control information flow from $h_{t-1}(q)$ to generate updated composite state $\tilde{C}(q)$.  This generated state is again gated by gate $l_2(q)$ which controls information flow from present with gated past state. Finally, the complement gate $(1-l_2(q))$ restores the past gated information.

\begin{equation} \label{eq}
\begin{split}
    C(q) & = [h_{t-1}(q), x_t(q)]\\
    l_1(q) & =\sigma(Conv1d(C(q)),  \\ 
    l_2(q) & =\sigma(Conv1d(C(q)),   \\ 
    \tilde{C}(q) &= \tanh(Conv1d([l_1(q) * h_{t-1}(q), x_t(q)]) \\
    h_t(q) & =(1-l_2(q))* h_{t-1}(q)+l_2(q) * \tilde{C}(q) \\
    h_t(q) & = DeformConv(h_t(q))
\end{split}    
\end{equation}

Following this, we process the non-overlapping regions by use of a deformable convolution \cite{zhu2019deformable} using a kernel of size $S$. Such an operation facilitates convolution at offset-defined locations (indicated by green arrow lines in right of Figure \ref{combine}). The regressed deformable offsets for each location, bridges the gaps between non-overlapping regions and facilitate the convolution and thus extract temporal features related to the global movement of the object.

 The operations in equations \ref{eq} (State Fusion in Figure \ref{block})  are performed for each of M latent queries to selectively obtain intermediate output for each object based on pixel-wise BEV features. The Fusion Output is generated by averaging all temporally fused gated states for each latent query.

\subsection{Block Aggregation}
With the above procedure, all latent queries should ideally fuse information about every object in the present state. However, this is a strong assumption, since not all latent queries may capture all relevant objects, and even individual latent queries may not correspond to a complete object, but rather to specific facets of the same object. In order to address this, it is necessary to utilise more than one fusion block in order to ensure sufficient flexibility enabling the capture of both complete objects and all objects over time.  Figure \ref{layer} illustrates the arrangement of the fusion layer for $N$ fusion blocks. Multiple fusion blocks (with shared memory) are used to ensure that all objects are captured from the sequence of input frames and memory over time. The outputs of each fusion block are interpolated and aggregated to produce the final generalized feature representation. These connections are designed to allow the gradient to flow directly to each of the $M$ latent queries of each fusion block. Since no auxiliary losses are used, this is an important step to ensure that the gradient is not diminished for all learnable latent queries, as it would be in a feed-forward style. 

\section{Experiments}

\begin{table*}[h]
\centering
\caption{Quantitative benchmark for class \textit{car} on the nuScenes dataset}
\begin{tabular}{lccccc}
\toprule
 \textbf{Object detector} & \textbf{Type} & \textbf{AP4.0}(\%) $\uparrow$ & \textbf{mAP} (\%) $\uparrow$ & \textbf{rel. mAP} $\uparrow$ \\

 \midrule
 GNN \cite{svenningsson2021radar} &  point-based  & 24.7 & 13.7  & -57.7\% \\
 PointPillars \cite{lang2019pointpillars} &  grid-based & 37.0 & 22.0  & -32\% \\
 RPFA-Net \cite{xu2021rpfa} &  grid-based  & 38.3 & 23.1  & -28.7\% \\
 KPConvPillars \cite{ulrich2022improved}  & hybrid  &42.2 & 26.2   &  -19\% \\
 CenterPoint Tf. \cite{Saini_2024_CVPR} & grid-based  &43.4 & 32.4   &  baseline \\
 AttentiveGRU (ours)  &  grid-based  & \textbf{46.7} & \textbf{36.9}  &\textbf{ +14\%} \\
 AttentiveGRU (ours) + \\ CenterPoint Tf. &  grid-based  & \textbf{50.4} & \textbf{39.2}  &\textbf{ +21\%} \\
\bottomrule
\end{tabular}
\label{nu}
\end{table*}
\subsection{Dataset}
Given the large domain gap in terms of sensor characteristics between different radar sensors and levels of radar data, we tested our model on two different datasets. The nuScenes dataset restricts the number of submissions for its test dataset, prompting us to perform our comprehensive comparison with all relevant works, reimplemented from the literature, on its official validation dataset, following the test approach in \cite{ulrich2022improved}. Furthermore, although our network is capable of detecting object types such as bikes and pedestrians, it is important to note that the current state-of-the-art for radar-only detectors on the nuScenes dataset predominantly focuses on evaluating models based on their performance in detecting the 'car' class. This has become a de facto standard evaluation approach, as reflected in the works \cite{ulrich2022improved,svenningsson2021radar,yang2018pixor,meyer2021graph,Saini_2024_CVPR}. Consequently, in line with this common practice in literature, our quantitative evaluation on the nuScenes dataset focuses primarily on the "car" class, which accounts for the majority of the bounding boxes in the dataset. In our work, we use standard validation and training split of public nuScene dataset.

In order to conduct a comprehensive ablation of our model, we utilize an additional proprietary radar dataset from \cite{braun2021quantification} having a higher complexity. This dataset features finer class separations (see Table \ref{class}) based on vehicle dynamics (moving and stationary), and vehicle sizes (regular to large) enabling a more rigorous evaluation of our model. Large vehicles include types such as trucks and buses, while regular vehicles include cars, auto-rickshaws, etc. Furthermore, this dataset includes challenging real-world scenarios, such as fully occluded objects that are invisible to both lidar and camera, but detectable in radar data through multipath propagation. In our work, for this additional dataset we use a total of 21766 scenes for training and 9294 scenes for the test set.

\begin{table}[h]
\centering
\fontsize{10}{12}\selectfont
\caption{Class Categories for additional radar dataset}
\begin{tabular}{c}
\toprule
Classes \\
\midrule
 Vehicle Moving (VM)\\
Vehicle Stationary (VS) \\
 Large Vehicle Moving (LVM) \\
 Large Vehicle Stationary (LVS)  \\
 Bikes  \\
 Pedestrians  \\
\bottomrule
\end{tabular}
\label{class}
\end{table}

\subsection{Implementation Details}
We trained all the networks presented here using the Adam optimizer with a learning rate of $1 \times 10\textsuperscript{-4}$ for twenty-nine epochs and a rate of $1 \times 10\textsuperscript{-5}$ for the last 30\textsuperscript{th} epoch using early stopping, with a batch size of $1$, and on an Nvidia 2080TI GPU. The models are fitted using the sigmoid focal-based loss \cite{lin2017focal} to account for class imbalance in the training dataset. For the bounding box regression, the smooth L1 loss is used. For the FPN feature extractor, a 3-level pyramid with a feature dimension size of $64$ is employed throughout the architecture, while the transformer block utilizes the default parameter values from \cite{Saini_2024_CVPR}. In the proposed fusion layer, the latent query dimension for each fusion block is set to $64$ with a total of $M=32$ latent queries. A kernel size $S=3$ is selected for deformable convolution in state integration module and input of time sequence length T=2 is used for nuScene dataset while T=12 is used for proprietary dataset. In Figure \ref{block}, the mean is taken before applying deformable convolution to the intermediate output to speed up computations for the nuScenes dataset, as it is highly sparse. In total, the fusion layer contains three fusion blocks, resulting in a total of $32\times 3=96$ latent queries.

\subsection{Results}

For performance comparison, we adhered to the evaluation strategies employed by prior state-of-the-art works in automotive radar research, utilizing the 
nuScenes dataset. In the presented comparison of the proposed model with other detectors, we use the quantitative analysis presented in \cite{Saini_2024_CVPR,ulrich2022improved}, which reimplements and evaluates models from the literature on the nuScenes dataset with radar data as exclusive input. We choose the current best approach of Centerpoint Transformer \cite{Saini_2024_CVPR} (for nuScenes in 2024 with training and inference using radar only modality)  as our baseline. Additionally, as described in Section \ref{sec:formatting}, we do not compare our model to the related  works by Pu Wang et al.  \cite{li2022exploiting, yataka2024radar, yataka2024sira} because they are designed for high-resolution radar images from advanced sensors and as noted in \cite{ yataka2024sira}, are unsuitable for the sparse data of the nuScenes data \cite{li2022exploiting}.

Table \ref{nu} shows the results of our experiments on nuScenes. Comparing the  mean average precision (average: AP4.0, AP2.0, AP1.0 and AP0.5), it can be observed that compared to the current state-of-the-art transformer based model \cite{Saini_2024_CVPR} 13.8\% improvement can be seen by integrating temporal relations in CNN based FPNs. This impact is further explored in the following ablation studies. Notably, the best performance is obtained by substituting our proposed AttentiveGRU layer directly into the transformer \cite{Saini_2024_CVPR} for temporal fusion, which brings additional 21\% improvement over the previous state-of-the-art approach.

\section{Ablation}

\subsection{Setup}
Unlike images, the non-deterministic (Figure \ref{radar}) and sparse (Figure \ref{preprocess}) nature of radar data, particularly with BEV images in polar coordinates (range-azimuth), makes it challenging to effectively illustrate the added value using only a few selected images. Following the common practice in radar literature  \cite{meyer2021graph, svenningsson2021radar}, we provide a rigorous ablation of our methods, beyond scalar metrics, by analyzing detailed precision-recall curves for different class categories. To this end, we perform the subsequent ablation studies on the proprietary dataset \cite{braun2021quantification}, which allows for rigorous evaluation of the models by featuring finer class separations. Furthermore, with such class categories, we analyze how concepts of global receptive fields and spatio-temporal fields influence the final detection performance for objects based on their type, size and dynamics.

\begin{figure}[htbp]
    \def\svgwidth{\linewidth}
    \centering
    \fontsize{6}{8}\selectfont
    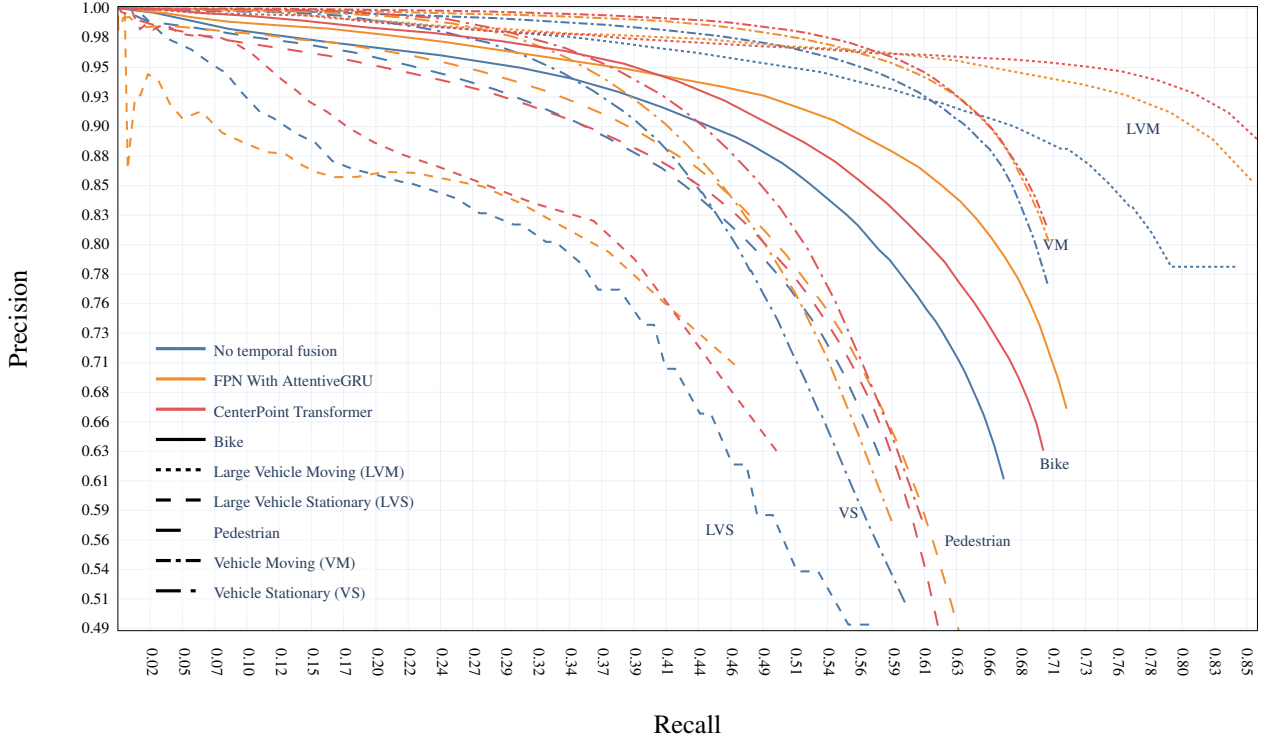
    \caption{Precision-Recall curves illustrating the detailed performance  of the proposed AttenitveGRU block}
    \label{pr}
\end{figure}

\subsection{Performance Curves}
Figure \ref{pr} compares the performance for the object detection task across all precision and recall metrics for bikes, pedestrians along-with large and regular classes of moving and stationary vehicles. Two of the curves in this figure shows performance of our FPN based model with and without AttentiveGRU module. Furthermore, a third curve shows performance of the current state-of-the-art model taken directly from \cite{Saini_2024_CVPR} without integrating it with the AttentiveGRU layer. 

\subsection{Discussion}
It can be observed that the addition of temporal fusion for the FPN based model brings a significant performance improvement for all classes under all operating conditions, indicated by a larger area under the curve (AUC). This improvement means that any chosen operating point will provide superior performance compared to alternatives, without requiring a trade-off between precision and recall metrics. In particular, the addition of AttentiveGRU results in a significant performance gain for the moving classes such as Bike, which have micro-Doppler signal \cite{belgiovane2017micro}, resulting in a unique latent query representation and better identification and fusion of gated states.

One of the advantage of using performance curves is that it can help to outline the role of global receptive field versus the temporal field for detecting objects of different class categories under different recall-precision tradeoffs. Considering the AUC metric for the curves of the CenterPoint transformer and FPN with AttentiveGRU, it can be observed that, across all operating conditions of the models, temporal fusion outperforms the (higher) global receptive field of transformers for regular-sized objects such as the 'bike' and 'pedestrian' classes. In contrast, bigger objects from LVM and LVS classes favors larger receptive field, while regular moving vehicle (VM) class benefits equally from both. For stationary vehicles, due to absence of dynamics, global receptive field is more favorable. In summary, the ablation studies highlight that temporal and global receptive fields offer distinct advantages for different object classes. Therefore, it is understandable that the model combining transformers with temporal fusion using AttentiveGRU delivers the best performance, as demonstrated in Table \ref{nu}.

\section{Summary}
In this paper, we have shown that the detection performance of objects of typical type, size and motion benefits from the introduction of spatio-temporal receptive fields in object detectors. We have presented an AttentiveGRU layer tailored to radar constraints, with an efficient attention gating module that dynamically identifies correlated object structures (pixel-wise BEV features) across radar scans for each object in the scene. These identified structures are fused with a state integration module to extract temporal features by processing overlapping and non-overlapping regions over time for each object. To make the fusion efficient while still being able to process challenging sparse inputs, we use recurrence together with an attention mechanism to fuse directly over sampled BEV pixel features. As a result, our approach, with linear complexity in space and time, demonstrates a significant performance improvement over existing radar-only object detection models for the challenging nuScenes dataset and an additional dataset, without reliance on external ego information or estimation effort. Furthermore, the modular design of the layer allows it to be easily incorporated into many radar or multimodal detection models. In our future work, we intend to enhance the multitasking capabilities of the architecture by incorporating complex tasks such as scene flow to improve the exploitation of correlated object structures beyond temporal feature extraction.

\bibliographystyle{unsrt}  
\bibliography{references}  

\end{document}